\documentclass[lettersize,journal]{IEEEtran}
\usepackage{amsmath,amsfonts}
\usepackage{algorithmic}
\usepackage{algorithm}
\usepackage{array}
\usepackage[caption=false,font=normalsize,labelfont=sf,textfont=sf]{subfig}
\usepackage{textcomp}
\usepackage{stfloats}
\usepackage{url}
\usepackage{verbatim}
\usepackage{graphicx}
\usepackage{cite}
\usepackage{booktabs}
\usepackage{multirow}
\usepackage{diagbox}
\usepackage{orcidlink}
\usepackage[table]{xcolor}
\begin{document}

\title{Autoregressive Multimodal Anomaly Detection with Large Language Models for Microservice Systems}

\author{Yixiang Tang\,\orcidlink{0009-0003-8603-9204}, Fagui Liu\,\orcidlink{0000-0003-1135-4982},~\IEEEmembership{Member,~IEEE}, Guoxiang Zhong\,\orcidlink{0000-0002-3998-7282}, Xi Yao\,\orcidlink{0009-0006-0204-9934}, C.L. Philip Chen\,\orcidlink{0000-0001-5451-7230},~\IEEEmembership{Fellow,~IEEE}
\thanks{This work was supported in part by the Guangdong Major Project of Basic and Applied Basic Research under Grant 2019B030302002, in part by the Science and Technology Major Project of Guangzhou under Grant 202007030006, in part by the Science and Technology Project of Guangdong Province under Grant 2021B1111600001, supported by the Major Key Project of PCL (Grant No. PCL2025AS13, PCL2024A05, PCL2025A08 and  PCL2025A13, PCL2025A11), in part by the National Natural Science Foundation of China under Grant U24B20151. (Corresponding authors: Fagui Liu) 
		
Yixiang Tang, Fagui Liu, Xi Yao, and C.L. Philip Chen are with the School of Computer Science and Engineering, South China University of Technology, Guangzhou 510006, China (e-mail: cstyx@mail.scut.edu.cn; fgliu@scut.edu.cn; csyaoxi@mail.scut.edu.cn; philip.chen@ieee.org).

Fagui Liu and Guoxiang Zhong are with the Department of New Networks, Peng Cheng Laboratory, Shenzhen 518000, China (e-mail: zhonggx01@pcl.ac.cn).}
\thanks{Manuscript received April 19, 2021; revised August 16, 2021.}}



\maketitle

\begin{abstract}
Microservice systems have become a dominant architecture for large-scale cloud applications, enabling rapid deployment and elastic scaling. However, ensuring the reliability and availability of microservice systems critically relies on effective anomaly detection, as failures and performance anomalies can quickly propagate across services and severely impact end-to-end system behavior. Existing multimodal anomaly detection approaches for microservice systems often struggle with two fundamental limitations: (i) the neglect of modality-specific characteristics in favor of cross-modality fusion, and (ii) the ``identity-shortcut" problem that affects the feature reconstruction process. To address these issues, we propose LLMMAD, an autoregressive multimodal anomaly detection framework for microservice systems based on Large Language Models (LLMs). Initially, to reconcile cross-modality fusion with modality-specific preservation, LLMMAD embeds and recombines multimodal data into a unified sequence for LLM processing to model inter-modality relationships, while employing modality-wise reconstruction heads to retain modality-specific information. Subsequently, to eliminate the identity-shortcut, we implement a prediction-driven autoregressive objective, under which the model predicts future states rather than copies current inputs. Finally, we conduct extensive experiments on three real-world datasets. The results demonstrate that LLMMAD consistently outperforms state-of-the-art methods, achieving F1-score improvements of 3.1\%, 1.4\%, and 0.9\%, respectively, thereby providing a more reliable solution for microservice monitoring.
\end{abstract}


\section{Introduction}
\label{sec:intro}
\IEEEPARstart{M}{icroservice} architecture has emerged as a mainstream paradigm for building large-scale enterprise applications, delivering enhanced flexibility, scalability, and fault isolation compared to monolithic systems~\cite{newman2021building}. In microservice architecture, a complex application is decomposed into a set of loosely coupled, fine-grained services, each running in its own process and communicating via lightweight mechanisms such as HTTP or message queues. This design enables independent deployment, rapid iteration, and elastic scaling, enhancing its industrial applicability.

However, the distributed and dynamic nature of microservice systems introduces significant challenges for ensuring reliability and operational safety. Failures can propagate across services in subtle ways, making timely and accurate anomaly detection critical yet difficult. Recognizing the critical role of service reliability, the leading cloud platforms such as eBay~\cite{xie2023unsupervised} and Microsoft Azure~\cite{zhang2021halo} have heavily invested in advancing anomaly detection technologies for microservice systems.

In microservice systems, system reliability engineers (SREs) collect metric, log, and trace data from running systems as operational data for monitoring~\cite{soldani2022anomaly}. Metrics are time-series data periodically
collected by monitoring systems, reflecting both application performance and resource states. Logs are sequences of semi-structured events recording service runtime behaviors. Traces consist of multiple spans, each representing a service
invocation and including details such as latency, status code, and operation type.

Current research has developed numerous anomaly detection algorithms for microservice systems. However, most existing methods rely primarily on single-modality data~\cite{meng2019loganomaly, liu2020unsupervised, tuli2022tranad, xie2023unsupervised}. Prior studies~\cite{lee2023eadro, ren2023triple} have shown that single-modality methods fail to capture anomaly signals associated with certain types of faults. This is because the information conveyed by a single modality is inherently limited and insufficient to represent the full system state. To overcome the inherent limitations of single-modal data, recent works~\cite{chen2023tracegra,zhang2022deeptralog,lee2023heterogeneous,wang2024madmm,lu2024misp,lee2023eadro,zhao2023robust,ren2023triple,huang2023twin,ding2025enhancing} have explored the use of multimodal data to achieve more accurate fault detection. These methods are shown in Table~\ref{tab:baseline}. While multimodal data provide richer information from multiple sources and offer a more comprehensive view of system health, current methods still exhibit several limitations:

\begin{table*}[t]
  \caption{A comprehensive comparison of various multimodal anomaly detection approaches of microservice systems. ``$\checkmark$" and ``$\times$" denote that the corresponding modality is used and not used, respectively.}
  \label{tab:baseline}
    \centering
    \setlength{\tabcolsep}{15pt}
    \resizebox{0.9\textwidth}{!}{
        \begin{tabular}{c ccc c c c}
        \hline
        \multirow{2}{*}{Approach} 
        & \multicolumn{3}{c}{Modality} 
        & \multirow{2}{*}{Fusion Method} 
        & \multirow{2}{*}{Temporal Extractor} 
        & \multirow{2}{*}{Learning Paradigm}\\
         & Metric & Log & Trace &  &  & \\
        \hline
        TraceGra~\cite{chen2023tracegra} & $\checkmark$ & $\times$ & $\checkmark$ & GNNs & LSTMs & Unsupervised\\
        DeepTraLog~\cite{zhang2022deeptralog} & $\times$ & $\checkmark$ & $\checkmark$ & GNNs & CNNs & Unsupervised\\
        Hades~\cite{lee2023heterogeneous} & $\checkmark$ & $\checkmark$ & $\times$ & Cross-Attention & CNNs+Transformers & Semi-Supervised\\
        MADMM~\cite{wang2024madmm} & $\checkmark$ & $\checkmark$ & $\times$ & Cross-Attention & GNNs+GRUs & Supervised\\
        MISP~\cite{lu2024misp} & $\checkmark$ & $\checkmark$ & $\times$ & Cross-Attention & Self-Attention & Supervised\\
        SCWarn~\cite{zhao2021identifying} & $\checkmark$ & $\checkmark$ & $\times$ & Concatenation & LSTMs & Unsupervised\\
        Eadro~\cite{lee2023eadro}   & $\checkmark$ & $\checkmark$ & $\checkmark$ & GNNs  & CNNs & Supervised\\
        AnoFusion~\cite{zhao2023robust}  & $\checkmark$ & $\checkmark$ & $\checkmark$ & GNNs & GRUs & Unsupervised\\
        Triple~\cite{ren2023triple}   & $\checkmark$  & $\checkmark$ & $\checkmark$ & GNNs & CNNs & Unsupervised\\
        MSTGAD~\cite{huang2023twin}  & $\checkmark$ & $\checkmark$ & $\checkmark$ & GNNs & CNNs & Semi-Supervised\\
        FUSION~\cite{ding2025enhancing}     & $\checkmark$ & $\checkmark$ & $\checkmark$ & Wavelet & CNNs & Unsupervised\\
        \hline
        LLMMAD     & $\checkmark$ & $\checkmark$ & $\checkmark$ & LLMs & LLMs & Unsupervised\\  
        \hline
        \end{tabular}
    }
\vspace{-10pt}
\end{table*}

\begin{itemize}
    \item \textbf{Limitation 1: Neglect of modality-specific information.} Current multimodal anomaly detection approaches primarily focus on modeling either cross-modal relationships at the expense of modality-specific information. GNN-based approaches~\cite{zhao2023robust, ren2023triple, huang2023twin, lee2023eadro, zhang2022deeptralog} learn graph representations that inherently entangle modalities, resulting in tightly coupled modality interactions. Other approaches~\cite{lee2023heterogeneous, wang2024madmm, lu2024misp, ding2025enhancing, zhao2021identifying} preserve separate modality embeddings but ultimately integrate them via fusion methods based on attention, wavelet or merely concatenation, still prioritizing cross-modal relationships. Consequently, these methods ignore the modeling of modality-specific information, potentially overlooking unique anomaly cues that are only evident within individual modalities.
    \item \textbf{Limitation 2: Susceptibility to identity-shortcut.} Existing multimodal anomaly detection methods typically rely on temporal feature extractors (e.g., temporal CNNs~\cite{lee2023eadro, ren2023triple, huang2023twin, zhang2022deeptralog, ding2025enhancing, lee2023heterogeneous}, LSTMs~\cite{chen2023tracegra, zhao2021identifying},  GRUs~\cite{wang2024madmm, zhao2023robust}, or self-attention~\cite{lu2024misp}) to learn temporal patterns. Among these methods, unsupervised approaches usually employ reconstruction architectures for anomaly scoring. However, reconstruction-based models~\cite{shi2021unsupervised, xia2020synthesize, pourreza2021g2d, you2022unified} that utilize the aforementioned temporal feature extractors are susceptible to the “identity shortcut’’ problem, where the decoder reproduces the input disregarding its semantics. This problem becomes more severe in multimodal settings, as the increased complexity of multimodal data makes it easier to exploit. As a result, the model fails to produce discriminative representations for abnormal patterns, ultimately undermining the robustness and effectiveness of anomaly detection.
\end{itemize}

To systematically address these identified limitations, we propose LLMMAD, an autoregressive multimodal framework that adopts Large language models (LLMs) for unsupervised anomaly detection in microservice systems. More specifically, LLMMAD converts the data of three modalities into a sequence of embeddings arranged in chronological order. After passing through a LoRA-adapted~\cite{hu2022lora} LLM with parameters frozen, embedding-level predictions are obtained, and then they are reconstructed into the original modalities. LLMMAD not only analyzes cross-modality relationships through unified multimodal LLM inference, but also captures modality-specific information via modality-wise reconstruction. Additionally, LLMMAD implements a prediction-driven autoregressive learning paradigm. By leveraging the LLM’s causal masking to infer future states from cross-modal contexts, LLMMAD decouples inputs from target predictions. This mechanism inherently eliminates the identity-shortcut and enhances sensitivity to both temporal and inter-modality anomalies. In addition, LLMMAD unifies decisions from modalities through the OR-logic aggregation method, providing fine-grained anomaly detection for microservice systems.

In this paper, we make the following contributions:

\begin{itemize}
    \item We propose LLMMAD, a LLM-based multimodal anomaly detection framework for microservice systems. LLMMAD jointly accounts for both cross-modality relationship and modality-specific information, addressing a key limitation of existing multimodal approaches.
    \item We design a prediction-driven autoregressive learning architecture for LLMMAD. This paradigm effectively prevents the “identity shortcut’’ problem that commonly plagues reconstruction-based anomaly detection methods, leading to more discriminative anomaly representations.
    \item We conduct comprehensive experiments on three real-world microservice datasets. The results demonstrate LLMMAD’s superiority over existing anomaly detection approaches for microservice systems.
\end{itemize}

\section{Related Work}
In this section, we provide an overview of previous works and discuss their limitations.


\subsection{Cross-Modality Knowledge Transfer via LLMs}
Cross-modality knowledge transfer denotes the process of aligning non-textual modalities into a language representation space. The knowledge learned from pretraining is shared to jointly model heterogeneous modalities and their interactions.

Recent studies have started to investigate whether LLMs can serve as a general backbone for cross-modal modeling of non-textual data. One-Fits-All~\cite{zhou2023one} extends this idea to time series data by leveraging frozen pretrained transformers to capture complex patterns in time series analysis. OneLLM~\cite{han2024onellm} further establishes a unified framework that aligns eight heterogeneous modalities into a language representation space through a universal projection module and a progressive alignment pipeline.

To the best of our knowledge, existing studies have not yet explored transforming multiple heterogeneous modalities in microservice systems into a unified language representation through cross-modal transfer. More importantly, leveraging LLMs for cross-modal transformation enables heterogeneous metrics, logs, and traces to be mapped into a shared sequential representation space with strong semantic and structural priors. This is crucial for modeling complex cross-modal dependencies and the downstream anomaly detection task.

\subsection{``Identity Shortcut" in Anomaly Detection}
Reconstruction-based anomaly detection relies on the assumption that models trained only on normal data can faithfully reconstruct normal patterns but fail on anomalous regions. However, such models often suffer from the ``identity shortcut" problem, where the reconstruction network optimized under reconstruction loss degenerates into an almost identity mapping and reconstructs both normal and anomalous samples equally well. You et al.~\cite{you2022unified} provide a systematic analysis of the ``identity shortcut" phenomenon and show that commonly used architectures (e.g., MLPs , CNNs, and Transformers) are inherently prone to converging to trivial identity solutions under reconstruction objectives. 

To mitigate the ``Identity Shortcut" issue, existing studies introduce additional constraints or training paradigms to regularize the reconstruction process. Structural and semantic instructional information~\cite{zhou2020encoding, shi2021unsupervised, xia2020synthesize} enforce high-level regularities in reconstruction. Memory-based mechanisms~\cite{gong2019memorizing, hou2021divide, park2020learning} constrain reconstruction to learned normal prototypes. Iterative refinement~\cite{dehaene2020iterative} progressively projects inputs toward the normal data manifold through multiple correction steps. Masking-based reconstruction~\cite{yan2021learning, fu2022mad, fang2024temporal} trains models to infer missing parts of the input. Pseudo-anomaly synthesis~\cite{pourreza2021g2d, collin2021improved, zavrtanik2021draem} augments normal data with synthetic anomalies and enforces recovery of clean samples.

Nevertheless, when LLMs are deployed as reconstructors, they remain susceptible to ``identity shortcut" due to their immense parameter space and powerful fitting capabilities. In contrast to existing reconstruction-based remedies, our work adopts a prediction-driven autoregressive learning paradigm, which explicitly decouples the input from the reconstruction target through causal modeling, thereby mitigating ``identity shortcut" in LLM-based anomaly detection.

\subsection{Anomaly Detection for Microservice Systems}
\subsubsection{Single-Modality Approaches}
Previous anomaly detection approaches are usually based
on logs~\cite{du2017deeplog, meng2019loganomaly, guo2021logbert, fu2023mlog} or metrics~\cite{zhong2023causeformer, tuli2022tranad, xu2021anomaly}, targeting traditional distributed systems without complex invocation relationships. DeepLog~\cite{du2017deeplog} models system logs as sequences of log keys and parameters, learns normal patterns with an LSTM, and detects anomalies via next-event prediction and parameter deviation. Donut~\cite{li2018robust} models seasonal KPI time series using a variational autoencoder with missing-data injection and detects anomalies based on reconstruction probability. Recently, some works~\cite{xie2023unsupervised, liu2020unsupervised, zhang2022putracead} have begun to use distributed traces for microservice anomaly detection. TraceVAE~\cite{xie2023unsupervised} employs a dual-variable graph variational autoencoder to encode trace structures and node execution times, and performs anomaly detection using entropy-gap–aware negative log-likelihood scoring.

Although effective within their respective modalities, these approaches can only detect anomalies that manifest in a single modality. Consequently, they fail to capture cross-modal correlations and cascading behaviors across modalities, which are essential for characterizing complex failure patterns.

\subsubsection{Multi-Modality Approaches}
Graph-based methods~\cite{chen2023tracegra, zhao2023robust, ren2023triple, huang2023twin, lee2023eadro, zhang2022deeptralog, he2026multimodal} organize multimodal data into unified graph structures. MSTGAD~\cite{huang2023twin} constructs a trace-centric graph backbone and incorporates metrics and logs as node-level features. AnoFusion~\cite{zhao2023robust} treats embedding channels from different modalities as graph nodes and estimates edge weights using mutual information. Other studies~\cite{lee2023heterogeneous, wang2024madmm, lu2024misp, ding2025enhancing, zhao2021identifying} adopt feature-level fusion strategies. Hades~\cite{lee2023heterogeneous} employs modality-specific feature extractors, Transformers for logs and CNNs for metrics, and subsequently fuses the extracted representations with cross-modality attention. FUSION~\cite{ding2025enhancing} fuses serialized multimodal data in the wavelet domain and performs probabilistic spatiotemporal modeling with a diffusion network for anomaly detection. 

However, a prevailing limitation across these methodologies is their singular focus on generating a holistic system representation for anomaly detection. By prioritizing an integrated global state, these models fail to sufficiently scrutinize the distinct information inherent to each modality. As a result, anomalies that manifest dominantly in a single modality or exhibit modality-specific signatures may be insufficiently represented in the fused features, which can ultimately limit detection robustness in complex microservice systems.

\section{Preliminary}
\subsection{Symbol Definition}

\begin{table}[t]
\centering
\renewcommand{\arraystretch}{1.5}
\caption{The symbols related to multimodal data and their descriptions.}
\label{tab:symbols}
\setlength{\tabcolsep}{15pt}
\resizebox{0.95\columnwidth}{!}{
    \begin{tabular}{cl}
    \hline
    \textbf{Symbols}  & \textbf{Definitions} \\
    \hline
    $\mathbf{X}_{\mathrm{metric}}$ & The time aligned metric matrix.\\
    $\mathbf{X}_{\mathrm{log}}$ & The time aligned log matrix.\\
    $\mathbf{X}_{\mathrm{trace}}$ & The time aligned trace matrix.\\
    $\mathbf{x}_t^\mathrm{metric}$ & The time aligned metric vector at timestep $t$.\\
    $\mathbf{x}_t^\mathrm{log}$ & The time aligned log vector at timestep $t$.\\
    $\mathbf{x}_t^\mathrm{trace}$ & The time aligned trace vector at timestep $t$.\\
    $\hat{\mathbf{x}}_t^\mathrm{metric}$ & The reconstructed metric vector at timestep $t$.\\
    $\hat{\mathbf{x}}_t^\mathrm{log}$ & The reconstructed log vector at timestep $t$.\\
    $\hat{\mathbf{x}}_t^\mathrm{trace}$ & The reconstructed trace vector at timestep $t$.\\
    $\mathbf{H}_\mathrm{emb}^\mathrm{metric}$ & The metric embedding sequence.\\
    $\mathbf{H}_\mathrm{emb}^\mathrm{log}$ & The log embedding sequence.\\
    $\mathbf{H}_\mathrm{emb}^\mathrm{trace}$ & The trace embedding sequence.\\
    $\mathbf{h}_{t}^\mathrm{metric}$ & The metric embedding vector at timestep $t$.\\
    $\mathbf{h}_{t}^\mathrm{log}$ & The log embedding vector at timestep $t$.\\
    $\mathbf{h}_{t}^\mathrm{trace}$ & The trace embedding vector at timestep $t$.\\
    $\hat{\mathbf{H}}_\mathrm{emb}^\mathrm{metric}$ & The predicted metric embedding sequence.\\
    $\hat{\mathbf{H}}_\mathrm{emb}^\mathrm{log}$ & The predicted log embedding sequence.\\
    $\hat{\mathbf{H}}_\mathrm{emb}^\mathrm{trace}$ & The predicted trace embedding sequence.\\
    $\hat{\mathbf{h}}_{t}^\mathrm{metric}$ & The predicted metric embedding vector at timestep $t$.\\
    $\hat{\mathbf{h}}_{t}^\mathrm{log}$ & The predicted log embedding vector at timestep $t$.\\
    $\hat{\mathbf{h}}_{t}^\mathrm{trace}$ & The predicted trace embedding vector at timestep $t$.\\
    $\hat{y}^{\mathrm{metric}}_t $ & The anomaly decision of metric modality at timestep $t$.\\
    $\hat{y}^{\mathrm{log}}_t $ & The anomaly decision of log modality at timestep $t$.\\
    $\hat{y}^{\mathrm{trace}}_t $ & The anomaly decision of trace modality at timestep $t$.\\
    $\mathcal{AS}_t^\mathrm{metric}$ & The anomaly score of metric modality at timestep $t$.\\
    $\mathcal{AS}_t^\mathrm{log}$ & The anomaly score of log modality at timestep $t$.\\
    $\mathcal{AS}_t^\mathrm{trace}$ & The anomaly score of trace modality at timestep $t$.\\
    \hline
    \end{tabular}
}
\vspace{-10pt}
\end{table}

Since this paper involves three different modalities, a relatively large number of symbols are used. To facilitate readability, the symbols related to multimodal data are summarized in Table~\ref{tab:symbols}. Other symbols are explained at the point of their first appearance in the text.

\subsection{Problem Formulation}
For a sliding time window $W=[t_1,t_2,\cdots,t_T]$, where $t_i$ presents the $i$th timestep of the window and $T$ is the number of timesteps in $W$, the timesteps are equally spaced along the temporal axis. Monitoring systems continuously or periodically collect multimodal operational data. The goal of anomaly detection is to identify anomalies occurring at each timestep by jointly leveraging information from all three types of data. 

In the following, we provide a formal description, including the mathematical formulation of each modality and the anomaly detection task within the window $W$.

\textbf{Metric.} We consider a set of instances in a microservice system, where each instance represents a container, a node, or a service. For the metric within $W$, the metrics of all instances form a set of multivariate time series as
\begin{equation}
\mathcal{M}=\left\{\{\mathbf{m}_t^i\}_{t=1}^{T} \;\middle|\; i=1,\ldots,N \right\}, 
\quad \mathbf{m}_t^i \in \mathbb{R}^{M},
\end{equation}
where $N$ is the number of instances, $M$ is the number of metrics and $\mathbf{m}_t^i$ is the metric vector of the $i$th instance at timestep $t$.

\textbf{Log.} A log entry consists of three core components, namely the timestep ($t$), severity level ($l$), and event description ($d$). A log sequence within window $W$ is represented as 

\begin{equation}
\mathcal{L} = \{\langle t_i, l_i, d_i \rangle\}_{i=1}^H,
\end{equation}
where $H$ indicates the total number of log entries in $W$ and $\langle t_i, l_i, d_i \rangle$ corresponds to the components of the $i$th log entry.

\textbf{Trace.} Each trace consists of multiple spans. A span represents an individual service invocation and is defined by several key attributes, including the caller service ($cr$), callee service ($ce$), duration ($d$), start time ($ts$), end time ($te$), operation name ($op$) and type ($ty$). For trace data collected within a time window W, the formal representation is given by
\begin{equation}  
\mathcal{S} = \{\langle cr_i, ce_i, d_i, ts_i, te_i, op_i, ty_i \rangle\}_{i=1}^K,  
\end{equation}  
where $K$ represents the total number of spans in W.
 
The task of anomaly detection is considered in a multimodal setting, where the objective is to learn a function $f$ that jointly integrates information from three modalities. This function is formulated as
\begin{equation}
    (y_1, y_2, \ldots, y_T) = f(\mathcal{M}, \mathcal{L}, \mathcal{S} \mid \theta),
\end{equation}
where $\theta$ denotes the model parameters, and $y_i \in \{0, 1\}$ represents the label at the $i$th timestep, with $y_i = 1$ indicating an anomaly and $y_i = 0$ indicating a normal state.

\section{Method}

\begin{figure}[t]
\centering
\includegraphics[width=0.95\columnwidth]{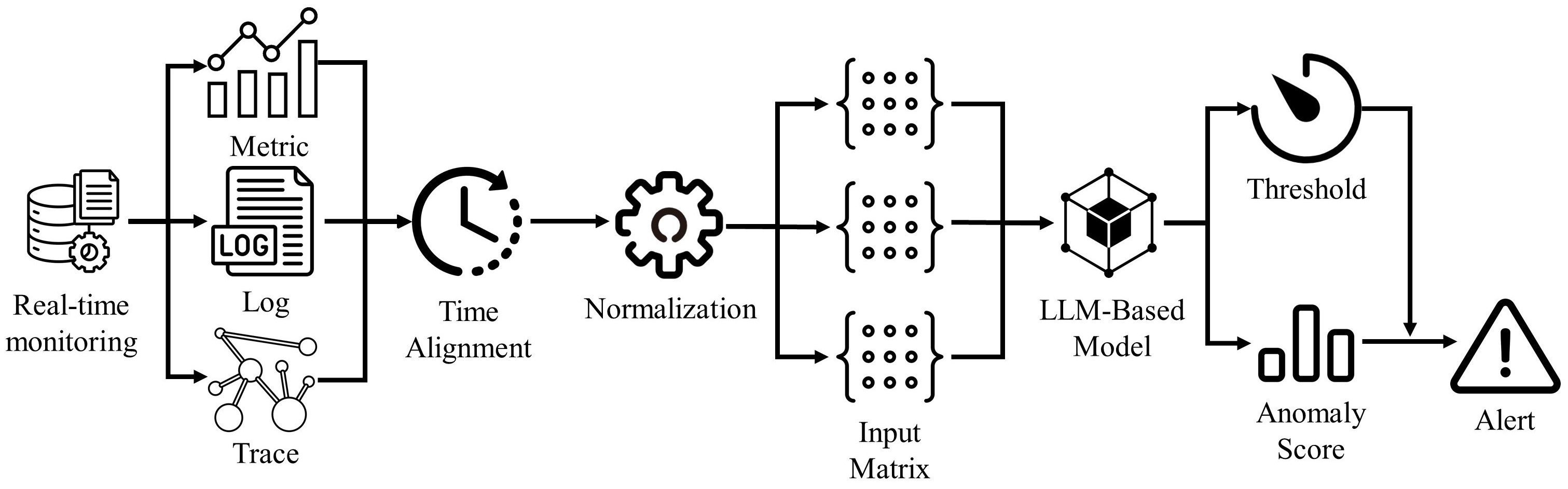} 
\caption{The overall pipeline of LLMMAD.}
\label{fig:framework}
\end{figure}

The framework of LLMMAD is presented in Fig.~\ref{fig:framework}. Multimodal monitoring data, including metrics, logs, and traces, are collected from the monitoring system. The unified input matrices are subsequently fed into the LLM-based model to obtain anomaly scores and threshold. For each timestep, alert is triggered when anomaly score exceed the threshold.

\subsection{Time Alignment}
For common metric collection systems, such as Prometheus\footnote{https://github.com/prometheus/prometheus}, metric data are typically sampled at fixed intervals (e.g., every 1\,s, 30\,s, or 1\,min). Based on these regularly sampled metric points, a sliding time window $W$ is defined. The timesteps within $W$ are aligned with the sampled data points, so that the metric sequences in each window share identical sampling intervals and have consistent lengths. In contrast, log data is event driven and log sequences vary in length while trace data consists of spans with variable number. To enable unified processing by the following model, log and trace sequences need to be aligned with the fixed temporal structure of metric data.

\begin{figure}[!t]
\centering
\includegraphics[width=0.9\columnwidth]{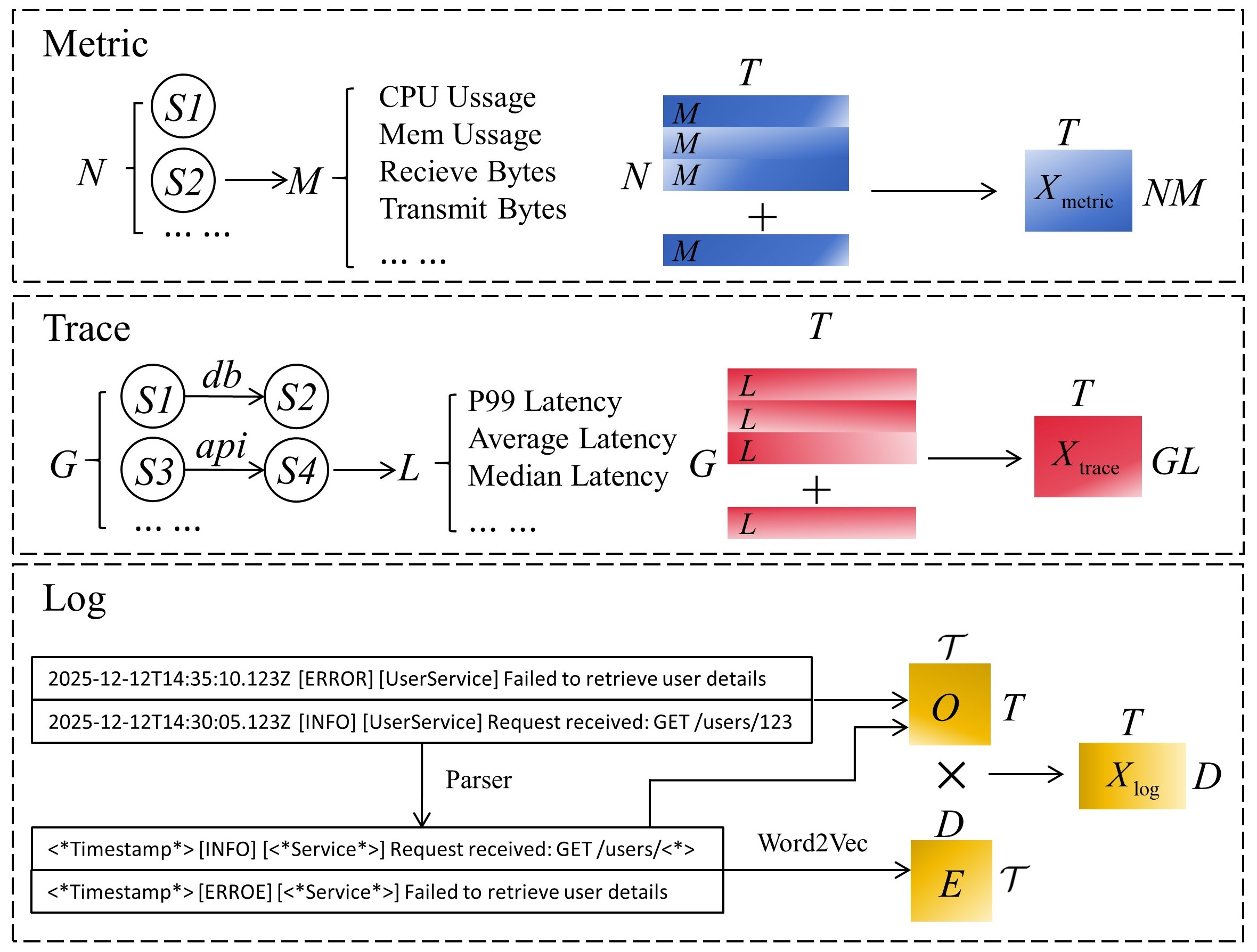}
\caption{The time alignment process of multimodal data.}
\label{fig:time_align}
\vspace{-10pt}
\end{figure}

In the following, we present the time alignment operation in detail. Additionally, Fig.~\ref{fig:time_align} is presented to visually illustrate the process of different modalities.

\subsubsection{Metric}

To aggregate the metrics from all instances, we concatenate their metric vectors along the metric dimension. An extended metric time series $\mathbf{X}_\mathrm{metric}$ is constructed as
\begin{equation}
\begin{aligned}
&\mathbf{X}_\mathrm{metric} = \{ \mathbf{m}_t \}_{t=1}^T \in \mathbb{R}^{T \times (N \cdot M)}, \\
&\mathbf{m}_t = \left[ \mathbf{m}_t^1 ; \mathbf{m}_t^2 ; \cdots ; \mathbf{m}_t^N \right] \in \mathbb{R}^{N \cdot M}, 
\mathbf{m}_t^{(i)} \in \mathbb{R}^{M}
\end{aligned}
\end{equation}
where $m_t$ is the concatenated metric vector of all instance at timestep $t$.

\subsubsection{Log}

The time alignment of log data starts by parsing raw log messages into structured templates. Based on the parsed templates, we represent their temporal occurrence using a binary log occurrence matrix as
\begin{equation}
\begin{aligned} 
    &\mathbf{O} = \{ \mathbf{o}_t \}_{t=1}^T \in \{0,1\}^{T{\times}\mathcal{T}},\\ 
    &\mathbf{o}_t = [\mathbf{o}_{t,1}, \mathbf{o}_{t,2}, \cdots, \mathbf{o}_{t,\mathcal{T}}] \in \{0,1\}^{\mathcal{T}},
\end{aligned}
\end{equation}
where $\mathcal{T}$ is the number of parsed log templates. Besides, $o_{t,\tau}$ is set to $1$ if a log corresponding to template $\tau$ appears at timestep $t$, and $0$ otherwise. This representation encodes the presence of log events over time.

To capture semantic information from log templates, each template $\tau$ is decomposed into a sequence of words as
\begin{equation}
\tau = \{w_1, ..., w_n\},
\end{equation}
where $w_i$ denotes the $i$th word of template $\tau$. A word corresponds to either an alphabetic string or a placeholder mask.

Each word is then mapped to a $D$-dimensional vector using Word2Vec~\cite{le2014distributed}. After word embedding, the template embedding is obtained via average pooling. This process is formulated as
\begin{equation}
\mathbf{e}_\tau = \frac{1}{|\tau|}\sum_{w_i \in \tau} \mathrm{Word2Vec}(w_i) \in \mathbb{R}^D,
\end{equation}
where $|\tau|$ is the number of words in the template $\tau$ and $\mathrm{Word2Vec}(\cdot)$ presents the Word2Vec function trained using the entire template corpus. Stacking all template embeddings, the log template embedding matrix $\mathbf{E}$ can be obtained as

\begin{equation}
    \mathbf{E} = \{ \mathbf{e}_{\tau} \}_{{\tau}=1}^\mathcal{T}\in \mathbb{R}^{\mathcal{T}{\times}D}, 
    \mathbf{e}_{\tau} \in \mathbb{R}^{D}.
\end{equation}

Finally, the semantic and occurrence information of log data are integrated by multiplying the log occurrence matrix and the template embedding matrix as
\begin{equation}
\mathbf{X}_{\mathrm{log}} = \mathbf{O} \times \mathbf{E} \in \mathbb{R}^{T \times D}.
\end{equation}
This operation produces an aligned log sequence, where each timestep is represented by a $D$-dimensional feature vector.
The resulting representation is independent of the size of template set and its expansion.

\subsubsection{Trace}
The span set $\mathcal{S}$ is transformed into a temporal matrix by aggregating the duration of caller-callee-operation group at each timestep. Multiple statistical characteristics of duration can be selected based on specific circumstances, such as the P99th percentile, P95th percentile, mean, median, and standard deviation. The time-aligned span matrix $\mathbf{X}_{trace}$ is constructed as
\begin{equation}
\mathbf{X}_{\mathrm{trace}} = \{ \mathbf{s}_t \}_{t=1}^T \in \mathbb{R}^{T \times (L{\cdot}G)},
\mathbf{s}_t \in \mathbb{R}^{L{\cdot}G},
\end{equation}
where $G$ is the number of caller-callee-operation groups, $L$ is the number of selected statistical characteristics and $\mathbf{s}_t$ represents the feature vector of timestep $t$.

\subsection{Normalization}

Through time alignment, we obtain the temporally aligned representations of multimodal operational data as
\begin{equation}
    \begin{aligned}
        \mathbf{X}_{\mathrm{metric}} & \in \mathbb{R}^{T \times (N \cdot M)}, \\
        \mathbf{X}_{\mathrm{trace}}  & \in \mathbb{R}^{T \times (L \cdot G)}, \\
        \mathbf{X}_{\mathrm{log}}    & \in \mathbb{R}^{T \times D},
    \end{aligned}
\end{equation}

Before being fed into the subsequent modules, these matrices must be normalized to ensure numerical stability and balanced feature contributions during model training. In the following, we use $\mathbf{X}$ to denote the matrix from each modality.


However, metric data in microservice systems exhibits intrinsic scale heterogeneity, making normalization non-trivial. In particular, metric data often suffer from severe magnitude discrepancies across different variables. To mitigate this issue, we adopt a multi-stage normalization strategy. For metric variables with excessively large magnitudes, we first apply logarithmic compression as
\begin{equation}
\mathbf{X}^{\mathrm{log}} = \log(1 + \mathbf{X}),
\end{equation}
which effectively reduces scale variance while preserving relative variations. Conversely, for metric variables with extremely small magnitudes, we rescale them by a constant factor $\alpha$ as
\begin{equation}
\mathbf{X}^{\mathrm{scale}} = \alpha \cdot \mathbf{X}.
\end{equation}

After scale adjustment, we perform variable-wise robust normalization for all three modalities. Specifically, each feature dimension is normalized as
\begin{equation}
\mathbf{X}^{\mathrm{norm}} = \frac{\mathbf{X} - \mathrm{median}(\mathbf{X})}
{\mathrm{IQR}(\mathbf{X})},
\end{equation}
where $\mathrm{median}(\cdot)$ denotes the median and $\mathrm{IQR}(\cdot)$ represents the interquartile range.
Compared to standard normalization, this robust formulation alleviates the influence of outliers and heavy-tailed distributions.

\subsection{Model Architecture}

\begin{figure}[!t]
\centering
\includegraphics[width=0.9\columnwidth]{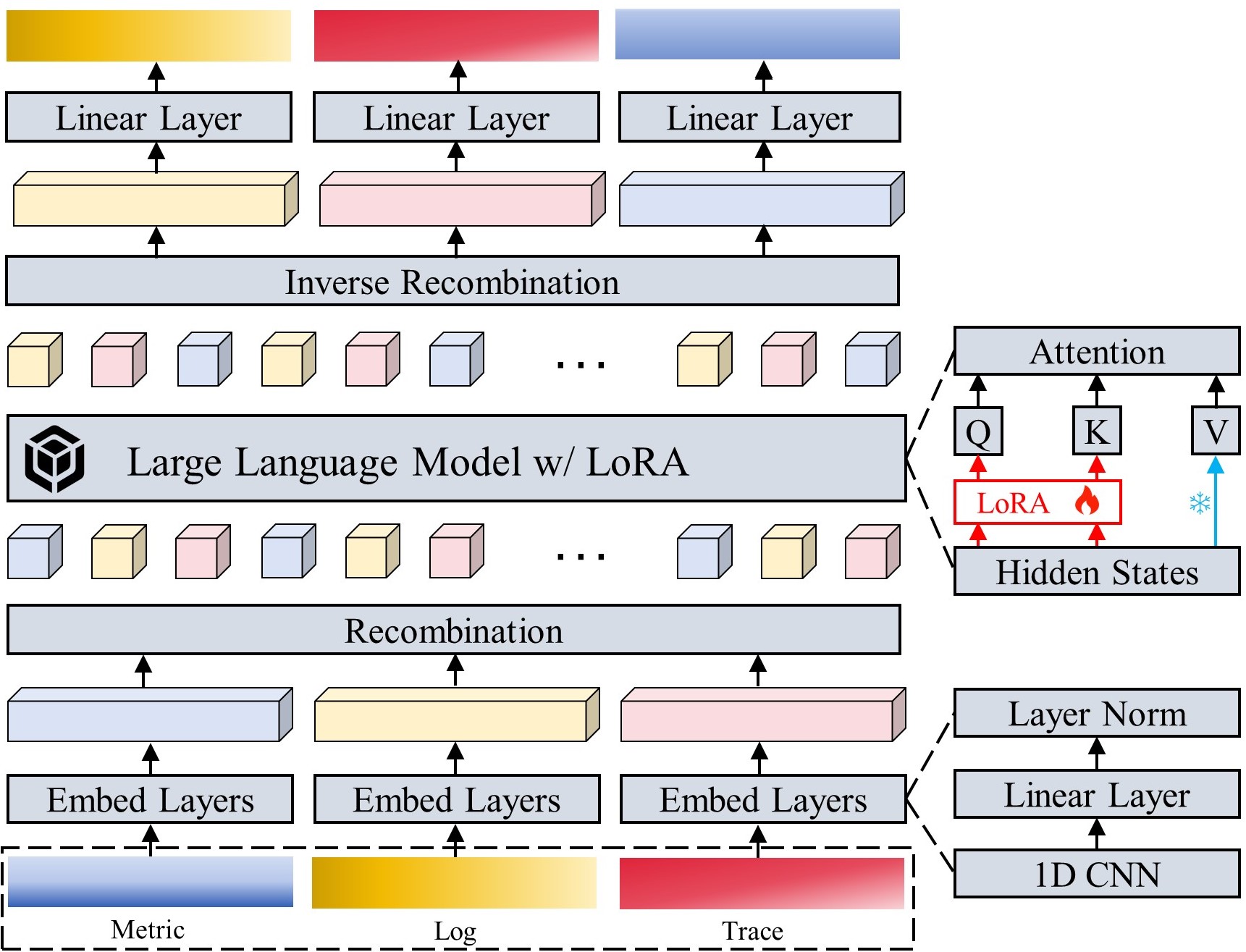}
\caption{The model architecture of LLMMAD. The light blue, light yellow, and light red blocks denote the intermediate feature matrices of metrics, logs, and traces, respectively.}
\label{fig:model}
\vspace{-10pt}
\end{figure}

This section presents the overall architecture of LLMMAD, and introduces each component in sequence to illustrate how the design supports effective anomaly detection (Fig.~\ref{fig:model}).

\subsubsection{Embedding Layers}
The embedding layers in our framework consist of a 1D CNN layer, a linear projection layer, and a layer normalization module. The 1D CNN layer is used to smooth temporal features by averaging short-term fluctuations and reducing high-frequency noise. This helps capture local sequential patterns while making the temporal features more stable. The linear layer then projects these smoothed features into a latent space to align their dimensionality with the LLM embeddings. Finally, layer normalization is applied to stabilize the feature distribution and enhance generalization, mitigating potential issues arising from feature scale variations across different modalities. Formally, the embedding process can be expressed as
\begin{equation}
\begin{aligned}
&\mathbf{H}_\mathrm{emb}^\mathrm{metric} = \mathrm{LayerNorm} \Big( \mathbf{W} \cdot \mathrm{Conv1D} (\mathbf{X}_\mathrm{metric}) \Big) \in \mathbb{R}^{T \times E},\\
&\mathbf{H}_\mathrm{emb}^\mathrm{log} = \mathrm{LayerNorm} \Big( \mathbf{W} \cdot \mathrm{Conv1D} (\mathbf{X}_\mathrm{log})\Big) \in \mathbb{R}^{T \times E},\\
&\mathbf{H}_\mathrm{emb}^\mathrm{trace} = \mathrm{LayerNorm} \Big( \mathbf{W} \cdot \mathrm{Conv1D} (\mathbf{X}_\mathrm{trace})\Big) \in \mathbb{R}^{T \times E},
\end{aligned}
\end{equation}
where $E$ is the embedding dimension of the selected LLM, $\mathbf{W}$ is the linear projection weights, $\mathrm{LayerNorm}$ is the fuction of layer normalization and $\mathrm{Conv1D}$ is the fuction of 1D CNN.

\subsubsection{LLM w/ LoRA}
In LLMMAD, the backbone LLM is employed with its original tokenizer, token embeddings, and prediction head removed while only retaining the position embeddings and transformer layers. All parameters of the LLM are frozen to preserve the knowledge acquired during pre-training. LoRA modules~\cite{hu2022lora} are introduced for efficient fine-tuning to achieve the bidirectional alignment of pre-trained semantic space with multimodal embeddings. In addition, this deisgn significantly reduces the training costs while maintaining generalization capabilities. Formally, the processing procedure is expressed as

\begin{equation}
\begin{aligned}
&\hat{\mathbf{H}} = \mathrm{LLM}_{\text{LoRA}}(\mathbf{H}),\\
&\hat{\mathbf{H}} \in \mathbb{R}^{3T \times E}, \mathbf{H} \in \mathbb{R}^{3T \times E},
\end{aligned}
\end{equation}
where $\hat{\mathbf{H}}$ is the last hidden state of the LLM. The definition and the detailed acquisition process of input $\mathbf{H}$ is presented in Section~\ref{recombine}.  Therefore, LLMMAD simultaneously extracts cross-modal information and temporal dependency through this component.

\subsubsection{Recombination and Inverse Recombination}
\label{recombine}
In the pipeline of LLMMAD, we introduce two novel operations, recombination and inverse recombination. They respectively perform temporal reorganization of the input and output embeddings of the LLM to support the fusion and separation of modal information.

Recombination is executed after the embedding layers and the multimodal embeddings are the inputs, which have the same shape $T\times E$. Specifically, the operation rearranges the embeddings by interleaving modalities along the temporal axis while preserving the original time order. Without loss of generality, we assume that the modalities are interleaved in the order of $metric,log,trace$, and this order is used consistently throughout the paper. The recombination operation itself is independent of the specific modality order. Formally, the recombination operation can be expressed as
\begin{equation}
\label{eq:recomb}
\begin{aligned}
\mathbf{H} =& \mathrm{Recomb}(\mathbf{H}_\mathrm{emb}^\mathrm{metric},\mathbf{H}_\mathrm{emb}^\mathrm{log},\mathbf{H}_\mathrm{emb}^\mathrm{trace})\\ =& [
\mathbf{h}_1^\mathrm{metric}, \mathbf{h}_1^\mathrm{log}, \mathbf{h}_1^\mathrm{trace}, \\
&\mathbf{h}_2^\mathrm{metric}, \mathbf{h}_2^\mathrm{log}, \mathbf{h}_2^\mathrm{trace}, \\
&\dots, \\
&\mathbf{h}_T^\mathrm{metric}, \mathbf{h}_T^\mathrm{log}, \mathbf{h}_T^\mathrm{trace}
]\in \mathbb{R}^{3T \times E}.
\end{aligned}
\end{equation}

The inverse recombination operation is executed after the LLM's inference and the last hidden state of the LLM is the input. As the inverse of the recombination operation, this operation disentangles the interleaved sequence by partitioning the embeddings back into their respective modality-specific streams. The inverse recombination is formulated as
\begin{equation}
\hat{\mathbf{H}}_\mathrm{emb}^\mathrm{log}, \hat{\mathbf{H}}_\mathrm{emb}^\mathrm{trace}, \hat{\mathbf{H}}_\mathrm{emb}^\mathrm{metric} = \mathrm{InverseRecomb}(\hat{\mathbf{H}}),
\label{eq:Inverserecomb}
\end{equation}
where the three output embeddings change the order to $log,trace,metric$ and share the shape $T\times E$ and are formulated as

\begin{equation}
\begin{aligned}
&\hat{\mathbf{H}}_\mathrm{emb}^\mathrm{log} = [\hat{\mathbf{h}}_1^\mathrm{log}, \hat{\mathbf{h}}_2^\mathrm{log}, \cdots, \hat{\mathbf{h}}_T^\mathrm{log}]\in \mathbb{R}^{T \times E},\\
&\hat{\mathbf{H}}_\mathrm{emb}^\mathrm{trace} = [\hat{\mathbf{h}}_1^\mathrm{trace}, \hat{\mathbf{h}}_{2}^\mathrm{trace}, \cdots, \hat{\mathbf{h}}_{T}^\mathrm{trace}]\in \mathbb{R}^{T \times E},\\
&\hat{\mathbf{H}}_\mathrm{emb}^\mathrm{metric} = [\hat{\mathbf{h}}_{1}^\mathrm{metric}, \hat{\mathbf{h}}_{2}^\mathrm{metric}, \cdots, \hat{\mathbf{h}}_{T}^\mathrm{metric}]\in \mathbb{R}^{T \times E}.\\ 
\end{aligned}
\end{equation}

It is worth noting that the modality order of the inverse recombination output is shifted by one position relative to the recombination input, as each output  is the prediction of the next timestep. Consequently, the order is $metric,log,trace$ in Formula (\ref{eq:recomb}) while $log,trace,metric$ in Formula (\ref{eq:Inverserecomb}).

In terms of function, the recombination facilitate the fusion of multimodal data while the inverse recombination assists the separation. In addition, these two operations also make multimodal data compatible with the autoregressive framework of LLMs.

\begin{figure}[!t]
\centering
\includegraphics[width=0.9\columnwidth]{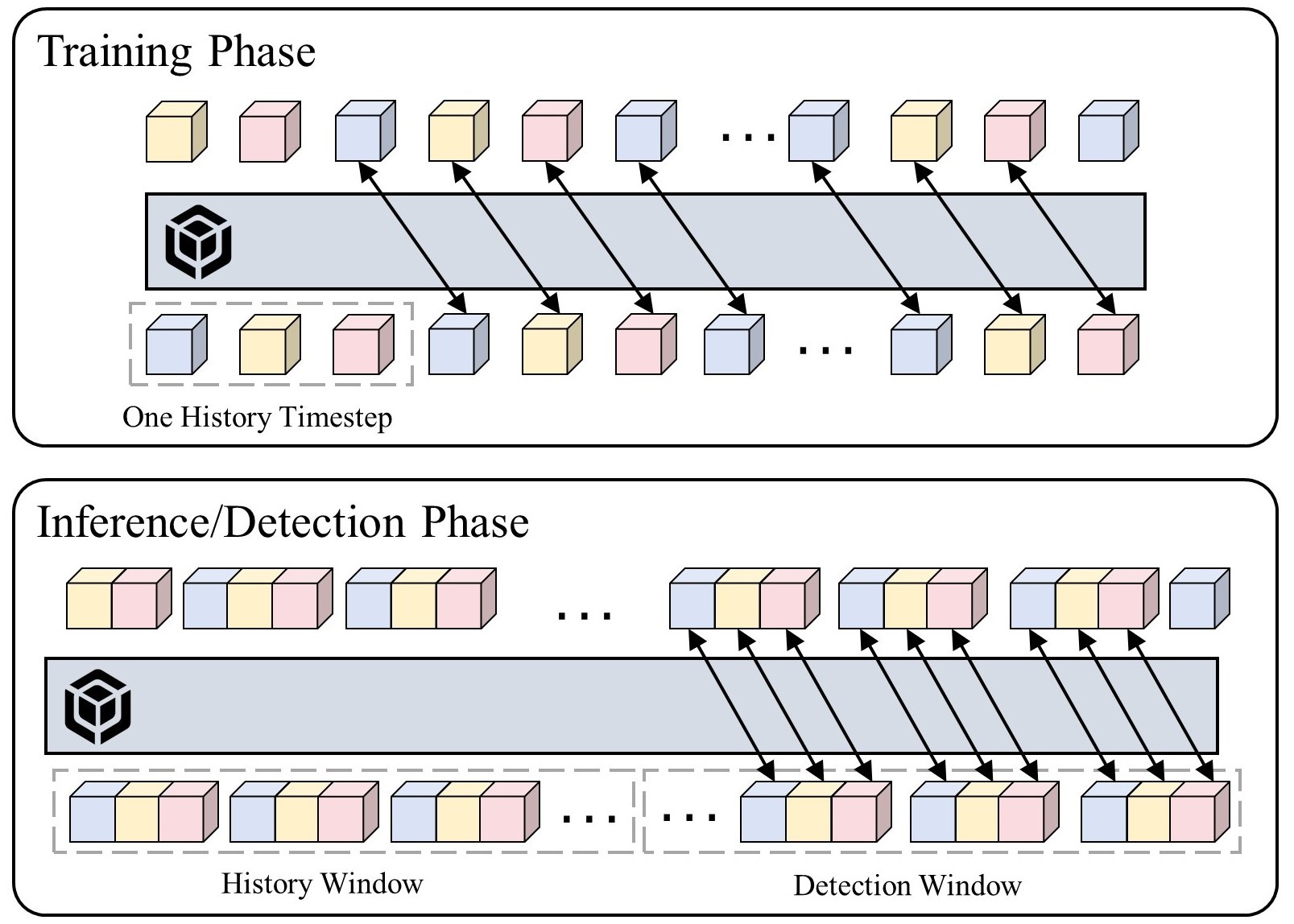}
\caption{The timestep correspondence of autogressive prediction. During training phase, every time step except the first one is predicted. During detection phase, only a fixed proportion of the latter part of the sliding window is predicted.}
\label{fig:train_and_infer}
\vspace{-10pt}
\end{figure}

\subsubsection{Modality-Wise Reconstruction}

The modality-wise reconstruction component serves as the final stage of LLMMAD. This component is designed to reconstruct the original data for each modality, ensuring the preservation of modality-specific features. By independently recovering the input data of each modality, it guarantees that the unique characteristics of each modality are maintained.

In particular, embeddings obtained through inverse recombination are passed through three independent linear layers for the reconstruction of each modality respectively. The modality-wise reconstruction process is formulated as

\begin{equation}
\begin{aligned}    
&\hat{\mathbf{X}}^\mathrm{metric} = \mathrm{Linear}(\hat{\mathbf{H}}_\mathrm{emb}^\mathrm{metric}) \in \mathbb{R}^{T \times (N \cdot M)}, \\
&\hat{\mathbf{X}}^\mathrm{log} = \mathrm{Linear}(\hat{\mathbf{H}}_\mathrm{emb}^\mathrm{log})  \in \mathbb{R}^{T \times D}, \\
&\hat{\mathbf{X}}^\mathrm{trace} = \mathrm{Linear}(\hat{\mathbf{H}}_\mathrm{emb}^\mathrm{trace})  \in \mathbb{R}^{T \times (L \cdot G)}.
\end{aligned}
\end{equation}

The reconstructed output serves as the final output of the entire model, and will be used for subsequent loss calculation and anomaly detection.

\subsection{Optimization Objective and Anomaly Inference}
\subsubsection{Prediction-Driven Autoregressive Learning}
During the training phase, we employ the prediction-driven autoregressive learning strategy, as predicting future observations encourages the model to capture meaningful temporal dependencies rather than merely reconstructing inputs.

To realize this prediction-driven paradigm, LLMMAD performs one-step-ahead prediction for each modality, where the reconstructed output at timestep $t$ is required to predict the input at timestep $t+1$. The one-step-ahead relationship between input and output is illustrated in the upper part of Fig.~\ref{fig:train_and_infer}, in which the first timestep of the sliding window does not contribute to the loss computation. The prediction objective is implemented using one-step-ahead MSE loss, and the individual losses for each modality are defined as follows:

\begin{equation}
\begin{aligned}
&\mathcal{L}_\mathrm{metric} = \frac{1}{T-1}  \sum_{t=1}^{T-1} 
\big\| \hat{\mathbf{x}}_t^\mathrm{metric} - \mathbf{x}_{t+1}^\mathrm{metric} \big\|_2^2,\\
&\mathcal{L}_\mathrm{log} = \frac{1}{T-1}  \sum_{t=2}^{T} 
\big\| \hat{\mathbf{x}}_t^\mathrm{log} - \mathbf{x}_{t+1}^\mathrm{log} \big\|_2^2,\\
&\mathcal{L}_\mathrm{trace} = \frac{1}{T-1}  \sum_{t=2}^{T} 
\big\| \hat{\mathbf{x}}_t^\mathrm{trace} - \mathbf{x}_{t+1}^\mathrm{trace} \big\|_2^2.
\end{aligned}
\end{equation}

The overall training loss $\mathcal{L}$ is defined as the sum of the losses from all three modalities and is formulated as

\begin{equation}
\mathcal{L} = \mathcal{L}_\mathrm{metric} + \mathcal{L}_\mathrm{log} + \mathcal{L}_\mathrm{trace}.
\end{equation}

This method helps avoid identity-shortcut behaviors by utilizing the causal masking mechanism in decoder-only LLMs, which ensures that the prediction of the timestep $t+1$ can only attend to timestep $t$ and the timestep before $t$. This restriction forces the model to rely on preceding context, preventing the model from simply copying input embeddings.

Moreover, this design aligns with the autoregressive training paradigm of decoder-only LLMs. Such consistency with the model’s learning mechanism reduces fine-tuning difficulty and overhead, facilitating effective adaptation to downstream modality-wise reconstruction.

\subsubsection{Anomaly Detection with OR-Logic Aggregation}
During the anomaly detection phase, we optimize the computation of the anomaly score. Due to the separate reconstruction of the three modalities, we introduce OR-logic aggregation to unify the final anomaly detection results.

The calculation of MSE loss of the detection phase is slightly different from the training phase. The specific difference is illustrated in Fig.~\ref{fig:train_and_infer}. In practice, the sliding window is divided into two parts according to a fixed proportion during detection phase, with the first half providing historical context and the second half used for anomaly detection. The original window $W$ was divided to $W_\mathrm{history}$ and $W_\mathrm{detection}$ as

\begin{equation}
\begin{aligned}
    &W_\mathrm{history} = \left[ t_1, t_2, \dots, t_{\left\lceil \gamma T \right\rceil} \right],\\ 
    &W_\mathrm{detection} = \left[ t_{\left\lceil \gamma T \right\rceil + 1}, 
    t_{\left\lceil \gamma T \right\rceil + 2}, \dots, t_T \right].
\end{aligned}
\end{equation}
where $\gamma \in (0,1)$ present the proportion of the length ratio of $W_\mathrm{history}$ to $W$ and is treated as a tunable hyperparameter. 

Anomaly scores are only calculated on $W_\mathrm{detection}$
, while $W_\mathrm{history}$ is not involved. The input of $W_\mathrm{history}$ serves to provide additional context for the model, helping it make more accurate predictions on the detection window. The specific calculation of anomaly score is as follows:


\begin{equation}
\begin{aligned}
&\mathcal{AS}_{t}^{\mathrm{metric}} = \big\| \hat{\mathbf{x}}_t^{\mathrm{metric}} - \mathbf{x}_{t+1}^{\mathrm{metric}} \big\|_2^2, \quad t = \lceil \gamma T \rceil, \dots, T-1; \\
&\mathcal{AS}_{t}^{\mathrm{log}} = \big\| \hat{\mathbf{x}}_t^{\mathrm{log}} - \mathbf{x}_{t+1}^{\mathrm{log}} \big\|_2^2, \quad t = \lceil \gamma T \rceil +1, \dots, T; \\
&\mathcal{AS}_{t}^{\mathrm{trace}} = \big\| \hat{\mathbf{x}}_t^{\mathrm{trace}} - \mathbf{x}_{t+1}^{\mathrm{trace}} \big\|_2^2, \quad t = \lceil \gamma T \rceil +1, \dots, T. 
\end{aligned}
\end{equation}

By selecting a threshold, the anomaly score is converted into a binary classification for each modality. A timestep is labeled as anomalous (1) if the score exceeds the threshold, and normal (0) otherwise. Thus, for the three modalities in each timestep, we obtain three binary anomaly judgments $\hat{y}^{\text{metric}}_t, 
\hat{y}^{\text{log}}_t, 
\hat{y}^{\text{trace}}_t$.

To obtain an overall anomaly decision across modalities, we apply an OR-logic aggregation strategy. Formally, the final anomaly label $\hat{y}_t$ obtained through OR-logic aggregation is formulated as

\begin{equation}
\begin{aligned}
&\hat{y}_t = 
\hat{y}^{\mathrm{metric}}_t 
\;\lor\;
\hat{y}^{\mathrm{log}}_t 
\;\lor\;
\hat{y}^{\mathrm{trace}}_t,\\
&\hat{y}_t,\hat{y}^{\mathrm{metric}}_t, 
\hat{y}^{\mathrm{log}}_t, 
\hat{y}^{\mathrm{trace}}_t \in \{0,1\}.
\label{eq:or_logic}
\end{aligned}
\end{equation}

In a nutshell, a timestep is considered anomalous if any of the three modalities is classified as anomalous. This simple yet effective aggregation strategy chooses to trust the judgment of each modality. 

\section{Experiments and Analysis}
In this section, we conduct comprehensive experiments to address the following research questions.
\begin{itemize}
    \item \textbf{RQ1}: How accurate is LLMMAD in detecting microservice anomalies?
    \item \textbf{RQ2}: How does each component and modality contribute to the effectiveness of LLMMAD?
    \item \textbf{RQ3}: How efficient is LLMMAD in terms of parameter scale and time overhead?
    \item \textbf{RQ4}: How effectively does LLMMAD mitigate the ``identity shortcut" problem?
    \item \textbf{RQ5}: How robust is LLMMAD to different LLM backbones and hyperparameter settings?
\end{itemize}

\subsection{Datasets}
We evaluate our method on three publicly available datasets collected from real-world microservice systems. Detailed descriptions of the datasets are as follows.

\begin{itemize}
    \item The SN dataset and the TT dataset~\cite{lee2023eadro}, are built upon two widely adopted microservice benchmarks. Specifically, the SN dataset is collected from the SocialNetwork system~\cite{gan2019open} and the TT dataset is derived from the TrainTicket system~\cite{zhou2018benchmarking}. Three kinds of faults, including CPU overload, network delay, and network packet loss, are manually injected into each microservice.
    \item The MSDS dataset~\cite{nedelkoski2020multi} originates from an OpenStack-based microservice cloud system and consists of two parts. The former is generated by executing workloads of sequential user requests, and the latter is generated by executing workloads of concurrent user requests. In our experiments, we use the part generated from concurrent requests. Three kinds of faults, including container restart faults, service API faults, and network agent disruptions, are injected into the Docker containers.
\end{itemize}

\begin{table}[t]
\centering
\caption{Statistics of the datasets used in our experiments, including time duration, instance numbers, metric types, log numbers and span numbers.}
\label{tab:dataset_statistics}
\setlength{\tabcolsep}{5pt}
    \resizebox{0.95\columnwidth}{!}{
        \begin{tabular}{lccccc}
        \toprule
        Dataset & Duration & Instance & Metric & Log & Span\\
        \midrule
        SN       & 9,870 s & 12 & 7 & 94,319 & 1,438,191\\
        TT       & 74,340 s & 27 & 7 & 65,616 &  158,238 \\
        MSDS     & 11,160 s & 5 & 5 & 130,326 &  1,514,358\\
        \bottomrule
        \end{tabular}
    }
\vspace{-10pt}
\end{table}

Table~\ref{tab:dataset_statistics} presents the characteristics of the datasets used in our experiments. These datasets exhibit substantial diversity in time duration, instance number, log, and span volume, covering a wide range of system behaviors and data distributions. These variability allows us to comprehensively evaluate the robustness and generalization ability of our method under different system scales and temporal characteristics.

\subsection{Experimental Settings}
All methods, including LLMMAD and the baselines, are implemented in PyTorch and evaluated on a Linux server equipped with eight NVIDIA Tesla V100 GPUs. Drain3~\cite{he2017drain} is used for log parsing, followed by Word2Vec trained with the skip-gram architecture, producing 64-dimensional embeddings. GPT-2 (117M)~\cite{radford2019language} is adopted as backbone LLM. All results are averaged over 20 runs. For anomaly detection, the threshold percentile is selected from $\{0.999, 0.995, 0.99, 0.98, 0.95, 0.90\}$ based on validation performance. Other default hyperparameters are set as follows.
\begin{itemize}
    \item Sliding window length: $T = 20$,
    \item Time interval: $t = 1\,\mathrm{s}$,
    \item Batch size: $B = 256$,
    \item Partition ratio: $\gamma = 0.5$,
    \item Epoch numbers: $epoch = 10$,
    \item Learning rate: $\eta = 0.01$.
\end{itemize}

All datasets are split into training, validation, and testing subsets. For SN and TT datasets with distinct normal and anomalous periods, only normal periods are used for training, while anomalous periods are split into validation and testing subsets with a ratio of 1:3. For MSDS dataset where anomalies occur throughout the entire timeline, the data are divided into training, validation, and testing sets with a ratio of 6:1:3, and anomalous samples are removed from the training subset.

\subsection{RQ1: How accurate is LLMMAD in detecting microservice anomalies?}

\begin{table*}[t]
    \centering
    \caption{Performance comparison between LLMMAD and baselines.
    The best, second-best, and third-best results in each column are marked by
    \colorbox[HTML]{F59194}{\phantom{\rule[-0.2ex]{1em}{1.4ex}}},
    \colorbox[HTML]{FAC791}{\phantom{\rule[-0.2ex]{1em}{1.4ex}}}, and
    \colorbox[HTML]{FFE699}{\phantom{\rule[-0.2ex]{1em}{1.4ex}}}, respectively.}
    \label{tab:comparison}
    \setlength{\tabcolsep}{15pt}
    \resizebox{0.9\textwidth}{!}{
        \begin{tabular}{l ccc ccc ccc}
        \toprule
        Dataset & \multicolumn{3}{c}{SN} & \multicolumn{3}{c}{TT} & \multicolumn{3}{c}{MSDS} \\
        \midrule
        Algorithm & P & R & F1 & P & R & F1 & P & R & F1 \\
        \midrule
        LogAnomaly & 0.751 & 0.678 & 0.713 & 0.612 & 0.796 & 0.692 & 0.615 & 0.684 & 0.648\\
        TraceAnomaly & 0.601 & 0.476 & 0.531 & 0.687 & 0.611 & 0.647 & 0.641 & 0.718 & 0.677\\
        AnomalyTransformer & 0.876 & 0.640 & 0.740 & 0.828 & 0.621 & 0.709 & 0.755 & 0.725 & 0.740\\
        DeepTraLog & 0.821 & 0.791 & 0.806 & 0.768 & 0.742 & 0.755 & 0.784 & 0.801 & 0.792 \\
        SCWarn     & 0.748 & 0.823 & 0.784 & 0.801 & 0.726 & 0.762 & 0.719 & 0.794 & 0.755 \\
        Triple     
        & \cellcolor[HTML]{FAC791}0.883 
        & \cellcolor[HTML]{FFE699}0.861 
        & \cellcolor[HTML]{FFE699}0.872 
        & \cellcolor[HTML]{FFE699}0.829 
        & \cellcolor[HTML]{FFE699}0.804 
        & \cellcolor[HTML]{FFE699}0.816 
        & \cellcolor[HTML]{FFE699}0.838 
        & \cellcolor[HTML]{FFE699}0.859 
        & \cellcolor[HTML]{FFE699}0.848 \\
        Fusion 
        & \cellcolor[HTML]{FFE699}0.879 
        & \cellcolor[HTML]{FAC791}0.897
        & \cellcolor[HTML]{FAC791}0.898
        & \cellcolor[HTML]{FAC791}0.855 
        & \cellcolor[HTML]{F59194}0.935
        & \cellcolor[HTML]{FAC791}0.893
        & \cellcolor[HTML]{FAC791}0.852 
        & \cellcolor[HTML]{F59194}0.901
        & \cellcolor[HTML]{FAC791}0.876\\
        \midrule
        LLMMAD 
        & \cellcolor[HTML]{F59194}0.921 
        & \cellcolor[HTML]{F59194}0.938 
        & \cellcolor[HTML]{F59194}0.929 
        & \cellcolor[HTML]{F59194}0.887 
        & \cellcolor[HTML]{FAC791}0.928 
        & \cellcolor[HTML]{F59194}0.907 
        & \cellcolor[HTML]{F59194}0.871 
        & \cellcolor[HTML]{FAC791}0.899 
        & \cellcolor[HTML]{F59194}0.885\\
        \bottomrule
        \end{tabular}
    }
\vspace{-10pt}
\end{table*}

To comprehensively assess the anomaly detection accuracy of LLMMAD in microservice systems, we compare it against a diverse set of unimodal and multimodal baselines. For the sake of fairness, we only select unsupervised methods as the baseline. The unimodal baselines consist of log-based method LogAnomaly~\cite{meng2019loganomaly}, metric-based method AnomalyTransformer~\cite{xu2021anomaly}, and trace-based method TraceAnomaly~\cite{liu2020unsupervised}. We also opt for several recent multimodal approaches SCWarn~\cite{zhao2021identifying}, Triple~\cite{ren2023triple}, Deeptralog~\cite{zhang2022deeptralog}, and FUSION~\cite{ding2025enhancing}that integrate two or more modalities of data. 

Table~\ref{tab:comparison} presents a comprehensive comparison across three standard metrics, including precision (P), recall (R), and F1-score (F1). Overall, LLMMAD achieves the best F1-score on all three datasets while maintaining a favorable balance between precision and recall, demonstrating its robust anomaly detection performance across diverse microservice environments. Compared with the previous state-of-the-art method FUSION, LLMMAD further yields F1-score improvements of 3.1\%, 1.4\%, and 0.9\% on the selected datasets, respectively. 

In addition, a clear performance hierarchy can be observed with respect to modality utilization. The three-modality methods (LLMAD, FUSION, Triple) consistently outperform those using two modalities, while two-modality methods (SCWarn, DeepTralog) in turn surpass single-modality approaches (LogAnomaly, TraceAnomaly, AnomalyTransformer). This trend holds across all datasets, highlighting the importance of comprehensive multi-modal information for accurately characterizing complex system behaviors and detecting anomalies. These results further validate the effectiveness of LLMMAD in fully exploiting complementary information from multiple modalities to achieve superior detection accuracy.

\subsection{RQ2: How does each component and modality contribute to the effectiveness of LLMMAD?}
\label{sec:RQ2}

\begin{table*}[ht]
\centering
\caption{Ablation study on LLMMAD across data modalities and components.}
\setlength{\tabcolsep}{15pt}
\resizebox{0.9\textwidth}{!}{
\begin{tabular}{l|ccc ccc ccc}
\toprule
Dataset & \multicolumn{3}{c}{SN} & \multicolumn{3}{c}{TT} & \multicolumn{3}{c}{MSDS} \\
\midrule
Ablation Setting & P & R & F1 & P & R & F1 & P & R & F1\\
\midrule
LLMMAD 
& 0.921 & 0.938 & 0.929 
& 0.887 & 0.928 & 0.907 
& 0.871 & 0.899 & 0.885 \\
\midrule
\multicolumn{10}{c}{Modality Ablation} \\
\midrule
w/o metric 
& \cellcolor[HTML]{FFE699}0.790 
& \cellcolor[HTML]{FAC791}0.821 
& \cellcolor[HTML]{FFE699}0.805 
& 0.776 
& \cellcolor[HTML]{FAC791}0.793 
& \cellcolor[HTML]{FAC791}0.793 
& \cellcolor[HTML]{FFE699}0.812 
& \cellcolor[HTML]{FFE699}0.791 
& \cellcolor[HTML]{FFE699}0.801\\
w/o log 
& \cellcolor[HTML]{F59194}0.857 
& \cellcolor[HTML]{F59194}0.871 
& \cellcolor[HTML]{F59194}0.864 
& \cellcolor[HTML]{FAC791}0.791 
& \cellcolor[HTML]{F59194}0.856 
& \cellcolor[HTML]{F59194}0.822 
& \cellcolor[HTML]{F59194}0.875 
& \cellcolor[HTML]{FAC791}0.821 
& \cellcolor[HTML]{FAC791}0.847\\
w/o trace 
& \cellcolor[HTML]{FAC791}0.825 
& \cellcolor[HTML]{FFE699}0.812 
& \cellcolor[HTML]{FAC791}0.818 
& \cellcolor[HTML]{F59194}0.817 
& 0.751 
& \cellcolor[HTML]{FFE699}0.783 
& \cellcolor[HTML]{FAC791}0.841 
& \cellcolor[HTML]{F59194}0.865 
& \cellcolor[HTML]{F59194}0.853\\
w/o metrics \& log 
& 0.702 & 0.755 & 0.728 
& 0.752 & 0.694 & 0.722 
& 0.669 & 0.682 & 0.675\\
w/o log \& trace 
& 0.752 & 0.802 & 0.776 
& \cellcolor[HTML]{FFE699}0.782 
& \cellcolor[HTML]{FFE699}0.778 
& 0.780 
& 0.791 & 0.727 & 0.758\\
w/o metrics \& log  
& 0.725 & 0.710 & 0.717 
& 0.726 & 0.687 & 0.706 
& 0.611 & 0.662 & 0.635\\
\midrule
\multicolumn{10}{c}{Component Ablation} \\
\midrule
w/o robust embedding 
& \cellcolor[HTML]{FAC791}0.898 
& \cellcolor[HTML]{F59194}0.937 
& \cellcolor[HTML]{F59194}0.917
& \cellcolor[HTML]{FFE699}0.852 
& \cellcolor[HTML]{F59194}0.919 
& \cellcolor[HTML]{F59194}0.884 
& \cellcolor[HTML]{FAC791}0.865 
& \cellcolor[HTML]{F59194}0.871 
& \cellcolor[HTML]{F59194}0.868\\
w/o prediction  
& 0.864 
& \cellcolor[HTML]{FAC791}0.892 
& \cellcolor[HTML]{FAC791}0.878 
& 0.826 
& \cellcolor[HTML]{FAC791}0.814 
& \cellcolor[HTML]{FFE699}0.820 
& \cellcolor[HTML]{FFE699}0.841 
& \cellcolor[HTML]{FAC791}0.859 
& \cellcolor[HTML]{FAC791}0.850\\
w/o LoRA  
& \cellcolor[HTML]{FFE699}0.891 
& \cellcolor[HTML]{FFE699}0.826 
& \cellcolor[HTML]{FFE699}0.857 
& \cellcolor[HTML]{FAC791}0.927 
& \cellcolor[HTML]{FFE699}0.772 
& \cellcolor[HTML]{FAC791}0.842 
& 0.824 
& \cellcolor[HTML]{FFE699}0.596 
& 0.692\\
w/o OR-logic aggregation 
& \cellcolor[HTML]{F59194}0.974 
& 0.652 
& 0.781 
& \cellcolor[HTML]{F59194}0.967 
& 0.695 
& 0.809 
& \cellcolor[HTML]{F59194}0.931 
& 0.557 
& \cellcolor[HTML]{FFE699}0.697\\
\bottomrule
\end{tabular}
}
\label{tab:ablation}
\vspace{-10pt}
\end{table*}

We conduct an ablation study to evaluate the contribution of different components in LLMMAD. Both modality ablations and component ablations are performed. The experimental results are summarized in Table~\ref{tab:ablation}. 

For modality ablation, we analyze the contribution of each input modality and any combination of them. Reducing a single modality leads to a noticeable decline in detection performance. Removing two modalities results in a further degradation, indicating that each modality provides complementary information. Moreover, across all datasets, excluding the metric modality or any modality combination that includes metric causes the most significant performance drop, highlighting the dominant role of metric data in anomaly analysis.

For component ablation, we investigate the effect of key architectural and training components by removing or modifying each component independently. The specific ablation settings are described as follows.
\begin{itemize}
    \item \textbf{w/o robust embedding}: Replace the original embedding modules with a single linear projection layer.
    \item \textbf{w/o LoRA}: Disable LoRA modules.
    \item \textbf{w/o prediction}: Remove the prediction-driven autogressive mechanism and directly align the model output with the input of the same timestep.
    \item \textbf{w/o OR-logic aggregation}: Replace the or-logic aggregation strategy with majority voting.
\end{itemize}

The robust embedding layers are introduced to improve the model’s robustness and generalization capability. As shown in the ablation results, removing this component slightly degrades performance across all datasets.

Removing the prediction-driven autoregressive mechanism leads to a noticeable performance degradation. This confirms that prediction-based supervision provides a stronger and more informative training signal than direct alignment, as it encourages the model to learn meaningful temporal regularities rather than trivial identity mappings. Its specific role in alleviating the ``identity shortcut" problem is further analyzed in Section~\ref{sec:RQ4}.

When LoRA is disabled, the backbone LLM remains fully frozen. Although the performance degrades, the model still exhibits non-trivial anomaly detection capability, indicating that pre-trained LLMs possess inherent temporal modeling ability. In this setting, precision is consistently higher than recall, suggesting that the frozen LLM can identify only a limited subset of salient anomalies. This comparison further demonstrates that introducing a small number of trainable parameters via LoRA is sufficient and necessary to effectively adapt LLMs to multimodal anomaly detection.

Replacing the OR-logic aggregation strategy with majority voting causes a dramatic decrease in recall. Although precision is slightly improved, the overall F1-score drops significantly, indicating that the voting-based scheme is overly conservative and leads to a large number of missed anomalies. This observation also suggests that failures in microservice systems often manifest in only a single modality, rather than simultaneously across multiple modalities, making the OR-logic aggregation strategy more effective in balancing false positives and false negatives. This conclusion further demonstrates the necessity of modeling modality-specific information.

\subsection{RQ3: How efficient is LLMMAD in terms of parameter scale and time overhead?}
Although LLMs provide strong representation capabilities, their large parameter scale raises concerns regarding efficiency in time-sensitive scenarios. We therefore analyze the efficiency of LLMMAD in terms of model size as well as training and inference time.

Parameter statistics are summarized in Table~\ref{tab:efficiency_params}. Due to different input dimensionalities, the numbers of parameters in the embedding and output layers vary slightly across datasets. The results reported in Table~\ref{tab:efficiency_params} are based on the SN dataset as a representative example. Overall, LLMMAD updates approximately 260K parameters, accounting for less than 0.5\% of the total model parameters.

\begin{table}[t]
\centering
\caption{Parameter statistics of LLMMAD, including total and trainable parameters of different model components.}
\label{tab:efficiency_params}
\setlength{\tabcolsep}{3pt}
\resizebox{0.9\columnwidth}{!}{
    \begin{tabular}{lccc}
    \toprule
    \textbf{Component} & \textbf{Total Params} & \textbf{Trainable Params} & \textbf{Trainable Ratio (\%)} \\
    \midrule
    Embedding Layers        & $\approx60 K$   & $\approx60 K$ & $100$ \\
    LLM Backbone (Frozen)   & $\approx82 M$   & $0$ & $0$ \\
    LoRA Adapters           & $\approx150 K$  & $\approx150 K$ & $100$ \\
    Linear Output Layer     & $\approx50 K$   & $\approx50 K$ & $100$ \\
    \midrule
    \textbf{Total}          & $\approx82.26 M$ & $\approx260 K$        & $\approx0.3$ \\
    \bottomrule
    \end{tabular}
}
\vspace{-10pt}
\end{table}

\begin{figure}[t]
\centering
\includegraphics[width=0.85\columnwidth]{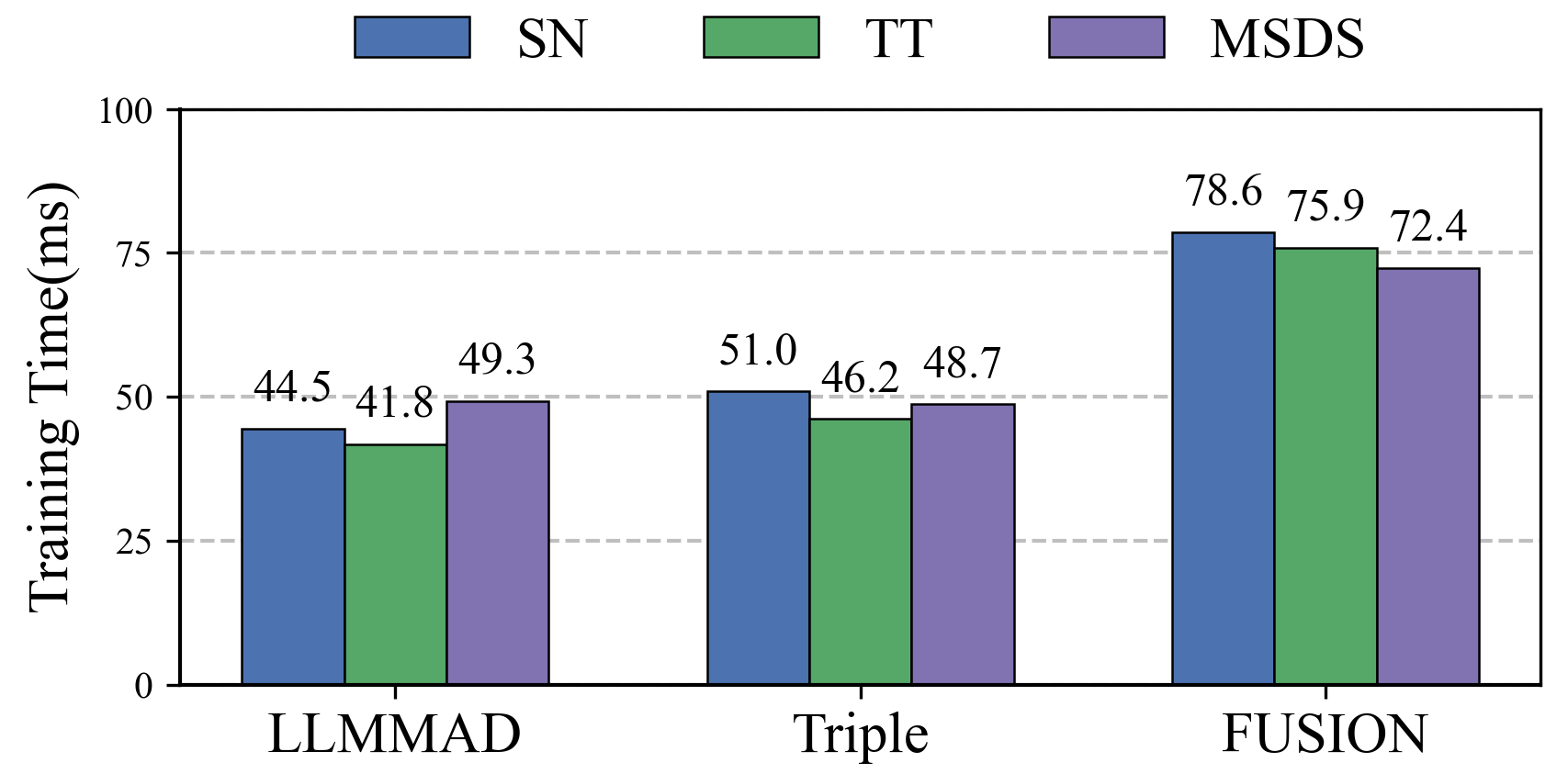}
\label{fig:training_time}
\caption{Average per-sample training time (in milliseconds) across datasets for the compared methods.}
\vspace{-10pt}
\end{figure}

\begin{figure}[t]
\centering
\includegraphics[width=0.85\columnwidth]
{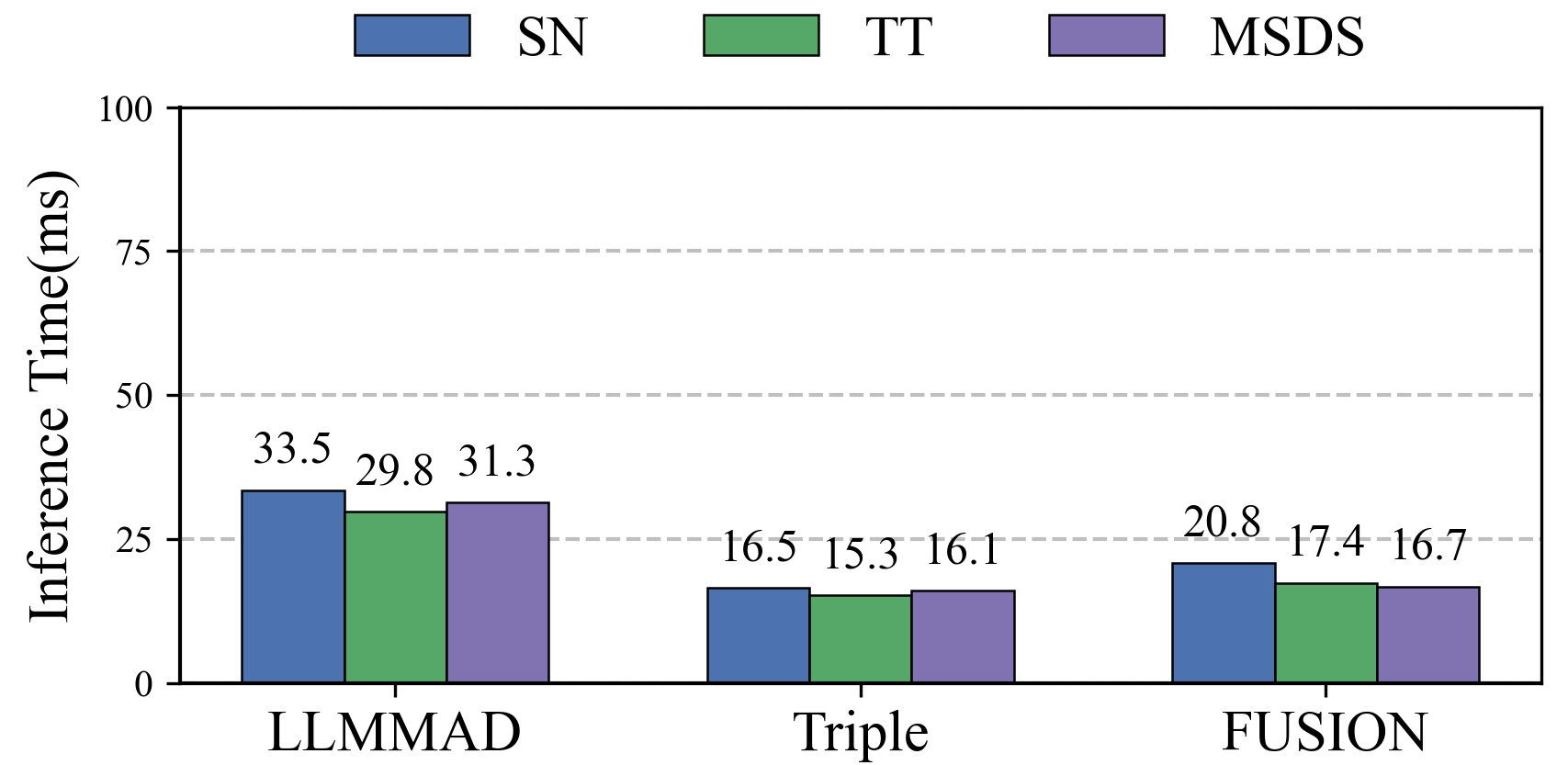}
\caption{Average per-sample inference time (in milliseconds) across datasets for the compared methods.}
\label{fig:inference_time}
\vspace{-10pt}
\end{figure}

\begin{figure*}[t]
\centering
\includegraphics[width=0.9\textwidth]{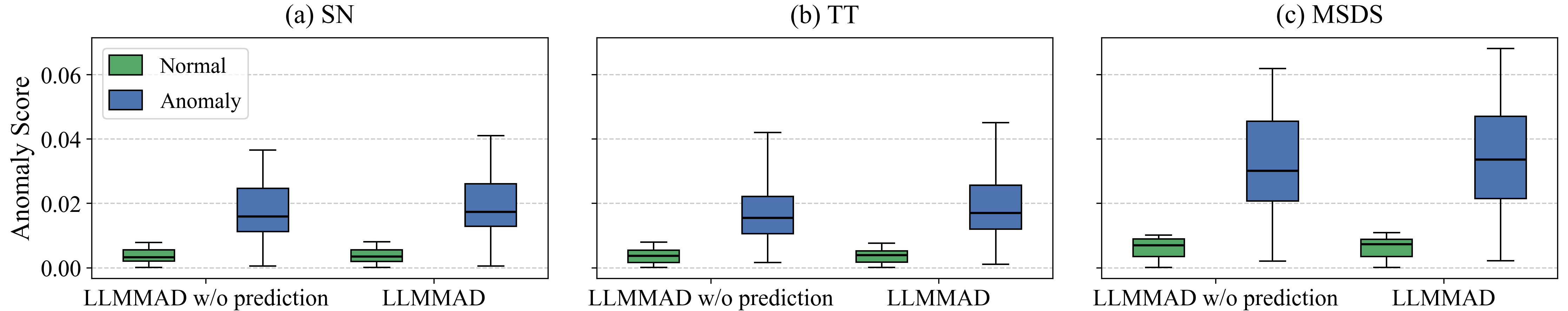}
\caption{Boxplots of anomaly scores for LLMMAD and LLMMAD w/o prediction on normal and anomalous samples.}
\label{fig:anomaly_boxplot}
\end{figure*}

\begin{table*}[t]
\centering
\caption{Performance comparison of LLMMAD with different underlying LLM backbones. Params denotes the number of model parameters, and Dim denotes the embedding dimension of the LLM.}
\setlength{\tabcolsep}{15pt}
\resizebox{0.9\textwidth}{!}{
\begin{tabular}{ccc|ccc ccc ccc}
\toprule
\multicolumn{3}{c|}{LLM} & \multicolumn{3}{c}{SN} & \multicolumn{3}{c}{TT} & \multicolumn{3}{c}{MSDS} \\
\midrule
Type & Params & Dim & P & R & F1 & P & R & F1 & P & R & F1 \\
\midrule
\multirow{3}{*}{GPT2}
& 117M & 768 
& \cellcolor[HTML]{FAC791}0.921 & 0.938 & \cellcolor[HTML]{F59194}0.929 
& \cellcolor[HTML]{FFE699}0.887 & 0.928 & \cellcolor[HTML]{FFE699}0.907 
& 0.871 & 0.899 & \cellcolor[HTML]{FAC791}0.885 \\
& 355M & 1024 
& 0.915 & 0.937 & \cellcolor[HTML]{FAC791}0.926 
& 0.879 & 0.931 & 0.904 
& \cellcolor[HTML]{FFE699}0.877 & 0.910 & \cellcolor[HTML]{F59194}0.893 \\
& 762M & 1280 
& 0.910 & 0.942 & \cellcolor[HTML]{FAC791}0.926 
& \cellcolor[HTML]{FAC791}0.889 & 0.927 & \cellcolor[HTML]{FAC791}0.908 
& 0.861 & 0.911 & \cellcolor[HTML]{FAC791}0.885 \\
\midrule
\multirow{2}{*}{Llama3.2}
& 1B & 2048 
& 0.866 & \cellcolor[HTML]{FFE699}0.956 & 0.909 
& 0.876 & 0.911 & 0.893 
& \cellcolor[HTML]{F59194}0.896 & 0.872 & \cellcolor[HTML]{FFE699}0.884 \\
& 3B & 3072 
& 0.850 & \cellcolor[HTML]{FAC791}0.960 & 0.902 
& 0.868 & 0.926 & 0.896 
& \cellcolor[HTML]{FAC791}0.892 & 0.879 & \cellcolor[HTML]{FAC791}0.885 \\
\midrule
\multirow{5}{*}{Qwen3}
& 0.6B & 1024 
& \cellcolor[HTML]{F59194}0.923 & 0.925 & \cellcolor[HTML]{FFE699}0.924 
& \cellcolor[HTML]{F59194}0.896 & 0.930 & \cellcolor[HTML]{F59194}0.913 
& 0.862 & 0.905 & 0.883 \\
& 1.7B & 2048 
& \cellcolor[HTML]{FFE699}0.920 & 0.923 & 0.921 
& \cellcolor[HTML]{FAC791}0.889 
& 0.922 & 0.896 & 0.865 
& \cellcolor[HTML]{FFE699}0.923 
& \cellcolor[HTML]{F59194}0.893 \\
& 4B & 2560 
& 0.869 & 0.933 & 0.900 
& 0.877 & \cellcolor[HTML]{FAC791}0.934 & 0.904 
& 0.823 & \cellcolor[HTML]{FAC791}0.933 & 0.875 \\
& 8B & 4096 
& 0.857 & 0.954 & 0.903 
& 0.884 
& \cellcolor[HTML]{FFE699}0.933 
& \cellcolor[HTML]{FAC791}0.908 
& 0.835 & 0.911 & 0.871 \\
& 32B & 5120 
& 0.813 & \cellcolor[HTML]{F59194}0.965 & 0.883 
& 0.851 & \cellcolor[HTML]{F59194}0.954 & 0.900 
& 0.812 & \cellcolor[HTML]{F59194}0.939 & 0.871 \\
\bottomrule
\end{tabular}
}
\label{tab:llm-ablation}
\vspace{-10pt}
\end{table*}

\begin{figure*}[t]
\centering
\includegraphics[width=0.9\textwidth]{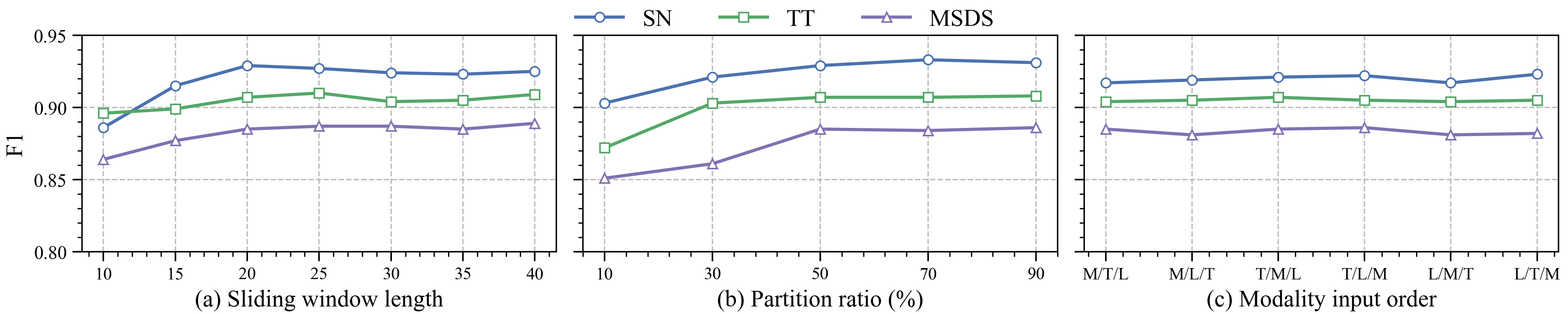}
\caption{The influence of three experimental setups on the performance of the method.}
\label{fig:parameter}
\vspace{-10pt}
\end{figure*}



Training and inference efficiency are evaluated using the per-sample training and inference time shown in Fig.~\ref{fig:training_time} and Fig.~\ref{fig:inference_time}, respectively, with comparisons against the multimodal baselines Triple and FUSION. LLMMAD achieves training efficiency comparable to Triple and is substantially faster than FUSION, while its inference time is only slightly higher than both baselines and remains sufficiently low for practical online anomaly detection. Moreover, unlike the baselines that typically exhibit a large gap between training and inference time, LLMMAD shows only a small difference between the two, indicating that the backpropagation overhead is relatively low. This behavior is mainly attributed to the small number of trainable parameters in LLMMAD, which reduces the cost of gradient computation.


\subsection{RQ4: How effectively does LLMMAD mitigate the ``identity shortcut" problem?}
\label{sec:RQ4}



To further investigate whether the prediction-driven mechanism helps mitigate the ``identity shortcut'' problem, we visualize the overall anomaly score (computed as the average anomaly scores of the three modalities) of LLMMAD and its ablated variant, LLMMAD w/o prediction, on both normal and anomalous samples across three datasets in Fig.~\ref{fig:anomaly_boxplot}.  For normal samples, the anomaly scores of LLMMAD and LLMMAD w/o prediction are generally similar across all datasets, indicating that removing prediction supervision has limited impact on normal data. In contrast, for anomalous samples, LLMMAD consistently assigns substantially higher anomaly scores compared to LLMMAD w/o prediction. Specifically, LLMMAD not only exhibits a higher median score but also shows elevated upper quartiles and maximum values, indicating a more pronounced separation between normal and anomalous points. Additionally, the overall spread of anomaly scores for LLMMAD on anomalous samples is larger, reflecting its enhanced sensitivity to abnormal patterns. These results supplement empirical evidence that the prediction-

\subsection{RQ5: How robust is LLMMAD to different LLM backbones and hyperparameter settings?}

To assess the robustness of LLMMAD under different configurations, this subsection conducts a two-fold analysis. We first study the effect of adopting different backbone LLMs and then analyze the sensitivity of LLMMAD to variations in key hyperparameters.

To isolate the influence of the backbone LLM, we compare LLMMAD using several LLM variants. Besides GPT-2 with 117M parameters, we also consider other GPT-2 variants of different sizes, as well as multiple versions of LLaMA3.2~\cite{dubey2024llama} and Qwen3~\cite{yang2025qwen3}, covering a wide range of parameter scales. For a fair comparison, the anomaly score threshold for each model is no longer selected based on the best performance, but is uniformly set to 0.99. 


The results of comparison between LLMs are summarized in Table~\ref{tab:llm-ablation}. It is observed that all evaluated LLMs exhibit a certain level of anomaly detection capability. Overall, the performance differences among LLMs of different types are marginal. Another notable observation is that for models of the same type but with different parameter scales, increasing the model size generally leads to higher Recall while slightly reducing Precision. This behavior is likely related to the increased embedding dimensionality associated with larger models. Higher-dimensional embeddings capture more fine-grained data characteristics, but also increase the risk of overfitting, thereby introducing more false positives. This issue is less evident on the TT dataset, which provides the largest amount of training data and thus better mitigates overfitting.



We next investigate the sensitivity of LLMMAD to key hyperparameters and experimental settings. The results are presented in Fig.~\ref{fig:parameter}. Specifically, we analyze the impact of the sliding window length $T$, the partition ratio $\gamma$, and the modality input order on performance.

Fig.~\ref{fig:parameter}(a) reports the impact of the sliding window length on LLMMAD. Overall, the model shows relatively stable performance when the window length is at least 20 across all three datasets. In contrast, a consistent and noticeable degradation is observed when the window length falls below this threshold, with the poorest performance at 10. This indicates that overly short temporal windows hinder the model’s ability to capture sufficiently complete temporal patterns, as critical long-range dependencies are truncated.

Fig.~\ref{fig:parameter}(b) shows the impact of the partition ratio $\gamma$ on detection performance. As $\gamma$ increases, the F1 scores on all three datasets improve, while the gains become marginal once $\gamma$ exceeds 50\%. A larger $\gamma$ incorporates a longer history for each detected timestep and thus better captures temporal dependencies, whereas a smaller $\gamma$ enables more timesteps to be reconstructed in parallel and yields higher practical efficiency. Therefore, $\gamma$ should be selected by balancing detection performance and efficiency.


Fig~\ref{fig:parameter}(c) evaluates the influence of modality input order on LLMMAD. Across all six possible permutations, the performance variations are negligible on all datasets. This observation demonstrates that LLMMAD is insensitive to modality input order, indicating that the recombination mechanism effectively integrates multimodal information without relying on a specific input order.

\section{Conclusion}

In this paper, we propose LLMMAD, an unsupervised framework that leverages LLMs for multimodal anomaly detection in microservice systems. To reconcile cross-modal interaction with modality-specific detail preservation, LLMMAD adopts a hybrid fusion-decoupling paradigm, in which multimodal embeddings are interleaved for unified LLM processing, while modality-wise reconstruction heads preserve modality-specific anomaly cues. Moreover, to address the ``identity shortcut" issue in reconstruction-based methods, LLMMAD employs a prediction-driven autoregressive objective that forces the model to infer future states from historical context rather than trivially copying inputs, thereby yielding more discriminative anomaly representations. Extensive experiments on three real-world datasets demonstrate that LLMMAD consistently outperforms state-of-the-art approaches. Overall, this work not only provides a reliable solution for automated microservice anomaly detection.

\bibliographystyle{IEEEtran}
\bibliography{refs}

\vfill

\end{document}